\documentclass[10pt,twocolumn,letterpaper]{article}

\usepackage{cvpr}
\usepackage{times}
\usepackage{epsfig}
\usepackage{graphicx}
\usepackage{amsmath}
\usepackage{amssymb}
\usepackage{caption}
\usepackage{floatrow}
\newfloatcommand{capbtabbox}{table}[][\FBwidth]

\usepackage[pagebackref=true,breaklinks=true,letterpaper=true,colorlinks,bookmarks=false]{hyperref}

 \cvprfinalcopy 

\def\cvprPaperID{2394} 

\ifcvprfinal\fi

\begin{document}

\title{Understanding learned CNN features through\\ Filter Decoding with Substitution\\}

\author{Ivet Rafegas\\
Computer Vision Center\\ C. Sc. Dpt. UAB. Bellaterra (Barcelona)\\
irafegas@cvc.uab.cat\\
\and
Maria Vanrell\\
Computer Vision Center \\ C. Sc. Dpt. UAB. Bellaterra (Barcelona)\\
maria@cvc.uab.cat\\
}

\maketitle

\begin{abstract}
   In parallel with the success of CNNs to solve vision problems, there is a growing interest in developing methodologies to understand and visualize the internal representations of these networks. How the responses of a trained CNN encode the visual information is a fundamental question both for computer and human vision research. Image representations provided by the first convolutional layer as well as the resolution change provided by the max-polling operation are easy to understand, however, as soon as a second and further convolutional layers are added in the representation, any intuition is lost. A usual way to deal with this problem has been to define deconvolutional networks that somehow allow to explore the internal representations of the most important activations towards the image space, where deconvolution is assumed as a convolution with the transposed filter. However, this assumption is not the best approximation of an inverse convolution. In this paper we propose a new assumption based on filter substitution to reverse the encoding of a convolutional layer. This provides us with a new tool to directly visualize any CNN single neuron as a filter in the first layer, this is in terms of the image space. 
\end{abstract}


\begin{figure}[t]
	\begin{center}
		\includegraphics[width=1\linewidth]{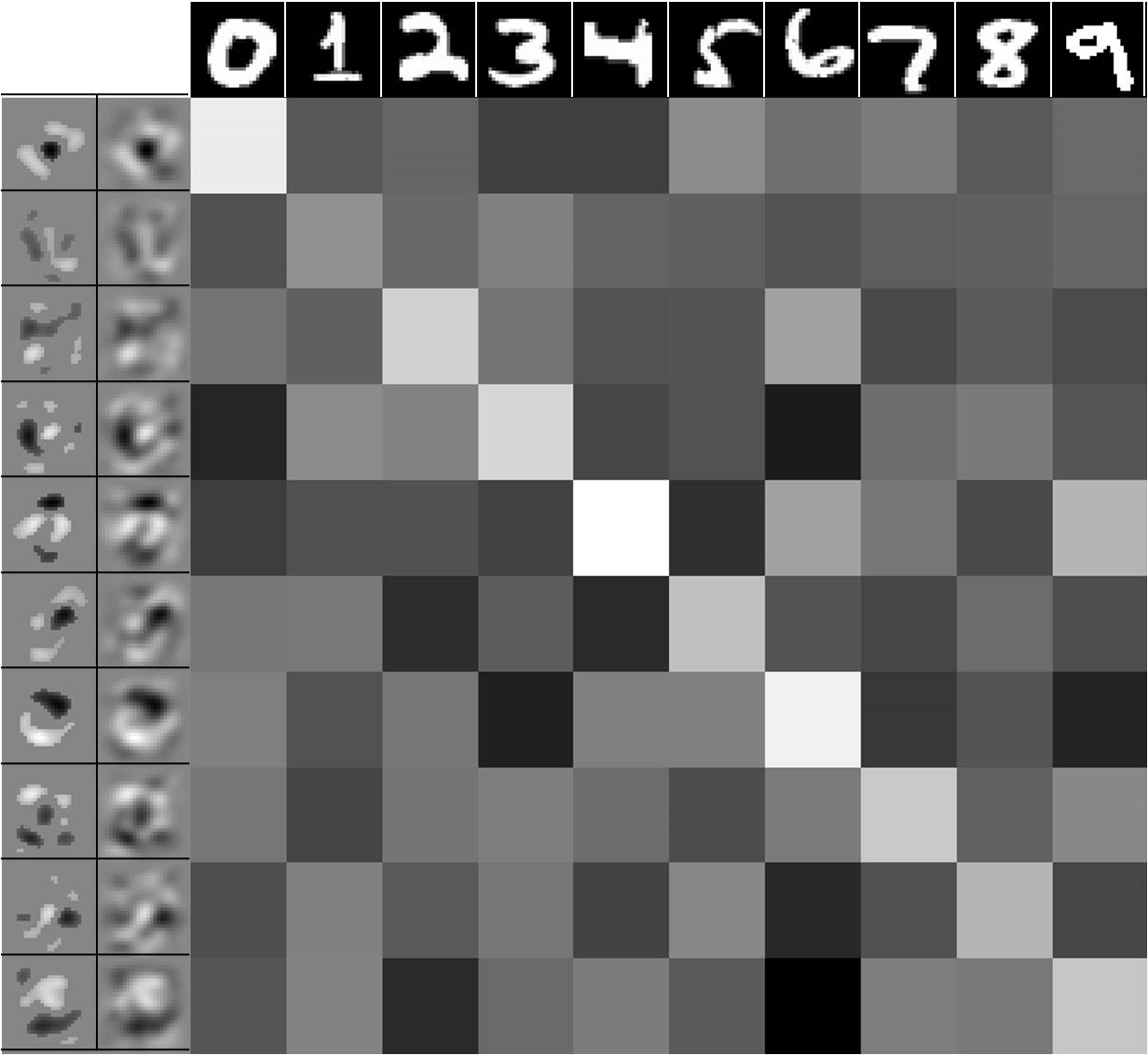}
	\end{center}
	\caption{MNIST images (top row) presenting maximum activation values by direct convolution with decoded filters (left columns) of LeNet highest layer. Left columns present two versions of the decoded filter, on the right the original one, on the left a threshold output to enhance filter shape. Activations are scalar values since images and filters have equal dimensions. Maximum responses at the diagonal  concludes a good match between decoded filters and class shape.}
	\label{fig:figure1}
\end{figure}

\section{Introduction}\label{sec:introduction}

The remarkable increase in performance of convolutional neural networks (CNNs) to solve computer vision problems is somehow diminished by the lack of understanding of the internal representations that capture the intrinsic image shapes. As we already mentioned before, the effects of the first convolutional layer in the net architecture can be easily understood. They act as feature detectors of specific spatial patterns that can be seen in the input image. Nevertheless, a second spatial convolution of these neuron activations entangles any intuition about the essential shape in the image. This problem is enlarged as we move forward in the hierarchy. On the other side, max pooling operators can be easily understood as simple image downsampling that just change the image resolution.

One of the most relevant works that tackles this understanding problem is the work of Zeiler and Fergus~\cite{Zeiler10} which is based on training a deconvolutional network.  Same authors in ~\cite{Zeiler11,Zeiler14} exploit the deconvolution operation to visualize the image feature projections with the highest activation at the corresponding layer. A deconvolutional network has also been used by Dosovitskiy and Brox~\cite{Dosovitskiy15b}. They use  a given representation provided by a CNN to estimate the deconvolutional network that minimizes the image reconstruction error for a set of images. On the contrary, Mahendrans and Vedaldi~\cite{Mahendran15} seek for the image whose representation minimizes the feature reconstruction error. Interestingly, object categorization using deconvolutional CNNs has been proved to be better than standard CNN when they are compared to electrophysiological data recorded during categorization tasks on primates. It has been shown by Cadieu \etal \cite{dicarlo14}. 

Other works that open the CNN black box, are those that try to understand how different scene variations are represented at the net levels. On one side the work of Aubry and Russell~\cite{Aubry15} the net is trained to encode a large synthetic dataset where the scene parameters such as 3D point of view, lighting conditions or object style have been gradually changed with a 3D CAD model. These images are represented with an standard CNN where the computed feature vectors can be qualitatively and quantitatively studied. Representation variations can be plotted in spaces of low dimensionality and can be used to predict similar variations in natural images. On another side, Dosovitskiy \etal \cite{Dosovitskiy15a} study natural scene variation represented with trained CNNs with the aim of generating accurate images of objects presenting unseen variations, such as, point of view, rotation or scale of the object. In this case, specific neurons can be studied as responsible of specific scene variations. 

In this work we present a new way to approach to the problem of visualizing internal representations, but focusing on understandig the filters themselves instead of the image responses. A similar idea has been outlined in a recent paper by Yosinski \etal \cite{Yosinski15} but with different assumptions. Our main contributions are twofold:

\begin{itemize}
	\item A new assumption to reverse the encoding of a convolutional layer is proposed. It is not based on the deconvolution as it has been used in previous works, but we propose a new assumption based on correlation as filter substitution. It allows to build a decoded version of the filters with a small errors and closer to the image space.
	
	\item An interpretation of decoded filters in terms of their intrinsic shapes instead of analyzing the image shape presenting the maximum activation. It allows to be applied to a wider range of trained CNNs. 
	
\end{itemize}

In the following section we firstly deal with the definition and implementation of the inverse convolution, this will bring us to hypothesize about a new assumption. Correlation assumption will provide us with a new method to backproject the filters towards the image space.

\section{Inverting convolution layers}\label{sec:inverting}

As we stated before, understanding CNN coding requires the projection of neural representations. To address this problem implies to invert the effects of layers. We want to note that we refer layers as single operators, either convolution or non-linearlity, usually max-pooling. Inverting the image coding through a convolutional layer was firstly approached by Kavukcuoglu \etal \cite{Kavukcuoglu10} and afterwards, the stacking of several layers was performed in \cite{Zeiler10}. In both works, deconvolution is approximated by the convolution with the transposed filter.

Considering that deconvolution refers to the inverse of the convolution, we want to make some considerations. The first one is that convolution with a transposed filter is by definition a correlation, which means to compute the similarity between the filter pattern at any image point. The second point is that the vast majority of computer vision libraries name the correlation implementation as convolution, e.g. MatConvNet \cite{vedaldi15matconvnet}. That make sense if we assume that in computer vision we pursuit to match specific templates in the images. Whatever is their implementation, in deconvolution nets it is assumed that correlation and convolution are inverse versions one each other, that is a rough approximation.

On the other hand, if we try to directly compute the inverse convolution of a given filter there are several ways to perform it. The most obvious approximation is to use the property that convolution becomes a product in the Fourier domain. Thus, the inverse filter can be computed as a direct division, which obviously not always exists due to zeros in the filter. There are several techniques to overcome this problem with error-minimization methods like Wiener filter. However, these techniques only reconstruct the original images for filters with specific constraints like narrow gaussian functions in image deblurring problems. Another approach is to linearize the convolution operation, which implies that the inverse filter can be computed by solving a system of linear equations. This problem is not always solvable for any filter since we can not assure that the rank of the coefficient matrix equals the one of the augmented matrix\footnote{This property is derived from the Rouché-Capelli theorem, where the augmented matrix of a linear equation system $Ax=b$ is $[A \quad b]$}. This bring us to a new interpretation of deconvolution that we hypothesize that each point at the convolution output represents the degree of similarity between a window surrounding the point and the filter pattern.
Heretofore, we assume that all the convolutional layers implement a correlation and we denote them with the  $\star$ symbol.

\begin{figure*}[t]
	\begin{center}
		\includegraphics[width=1\linewidth]{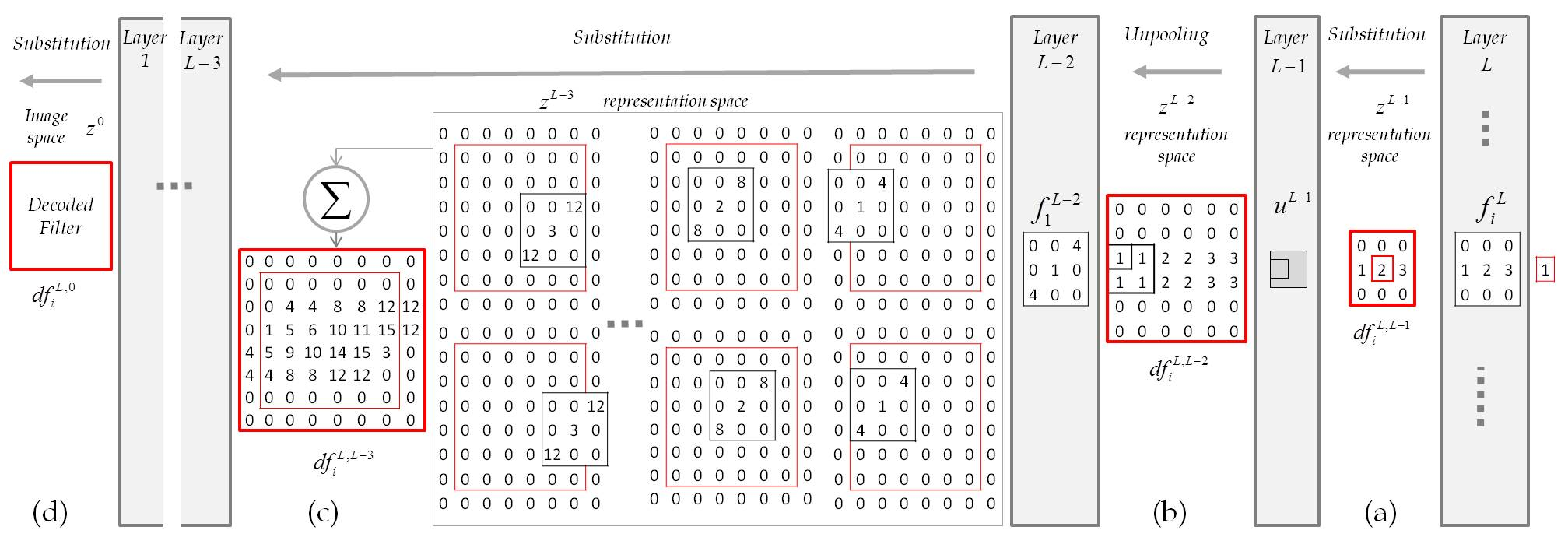}
	\end{center}
	\caption{Decoding scheme for filter $f_i^L$ (1-channel filter for simplicity). (a) 1st substitution step: a single pixel is substituted by a filter of layer $L$, is the initial case, $df_i^{L,L-1}=f_i^L$. (b) Unpooling step: doubles size of $df_i{L,L-1}$, from $3\times3$ to $6\times6$ giving $df_i^{L,L-2}$. (c) 2nd substitution step: filter on layer $L-2$ is substituted at each nonzero value of $df_i^{L,L-2}$ being multiplied by the pixel value, substitutions are added and denoted as $df_i^{L,L-3}$. (d) Decoded filter is denoted as $df_i^{N,0}$. }
	\label{fig:substitutionscheme}
\end{figure*}

\section{Filter decoding}\label{sec:filterdecoding}

In previous section we have seen the problems of inverting a convolutional layer of any network. To overcome this problem we propose an assumption about the effects of a convolutional layer when it is built in a hirarchical architecture. 

Our statement is that the value of a pixel $(x,y)$ belonging to the output of the $L$-convolutional layer, denoted as $z^L(x,y)$, can be understood as the shape similarity between the pixel region and the connected filter in layer $L$, which is denoted as $f^L$. This assumption can be applied to the representation of the same image in the inferior layer $L-1$ by substituting the filter shape onto the image $z^{L-1}(x,y)$. A strong activation in $z^L(x,y)$ is caused by a good match between the shape at  $z^{L-1}(x,y)$ and the filter $f^L$. We implement this idea by the substitution of the filter onto the pixel neighborhood. This procedure implies an increase in the image size as we descend in the network hierarchy. Our estimate of the image representation is given by

\begin{equation}
\hat{z}^{L-1}(x,y)=subst(z^L(x,y),f^L)
\end{equation}

The function $subst$ implements the assumption explained before. This idea also holds when applied directly onto the filters, since they are representing the substitution of a single pixel in the representation space on top. We denote as $df^{L,l}_{i,j}$ a decoded version of a filter $i$, channel $j$, at layer $L$. That can be computed as:

\begin{equation}
df^{L,l}_{j_i,c}:=subst(df^{L,l}_{i,j},f^{l}_{j,c})
\end{equation}
where $f^{l}_{j,c}$ is the channel $c$ of the filter $j$ in layer $L$. The size of the output image will be the size of the input plus the size of the filter on the corresponding dimensions. Note that when this function is applied onto an image neighbourhood, substitutions produce an overlaped region which share some pixels. We treat this cases with an addittion operation. Figure \ref{fig:substitutionscheme} illustrates a simple example of this function applied to an entire filter image.

In this way, we can define the filter decoding algorithm as a procedure to build an estimate of the intrinsic shape in the image space that is going to correlate with the filter at a layer $L$. 

For a given CNN architecture of $N$ layers the algorithm has the following steps:

\emph{
	\begin{center}
		\begin{tabular}{l} \hline
			filter\_decoding(L,k,CNN) \\  
			\hline
			\vspace{-0,25cm}\\
			$df^{L,L-1}_{k}:=f^L_k$  \\
			for $l=L-1:1$    \\
			\hspace{0,20cm} for $i=1:\#df^{l+1,l}$    \\
			\hspace{0,40cm} for $j=1:\#f^{l}$ \\
			\hspace{0,60cm} for $c=1:\#f^{l-1}$ \\
			\hspace{0,80cm}  if Layer $l$ is convolutional  \\
			\hspace{1cm}		$df^{L,l-1}_{j_i,c}:=subst(df^{L,l}_{i,j},f^{l}_{j,c})$ \\
			\hspace{0,80cm}  elseif Layer $l$ is pooling   \\
			\hspace{1cm}  	    	$df^{L,l-1}_{j_i,c}:=unpool(df^{L,l}_{i,j})$ \\
			\hspace{0,80cm}  endif \\
			end all for \\
			$df_k^{N,0}:= \sum_h \sum_c df^{L,l-1}_{h,c}$\\
			return($df^{N,0}_k$) \\
			\hline
		\end{tabular}
	\end{center}
}

\noindent The function $unpool$ performs a simple image upsampling. The input of the decoding function is the filter we want to decode, which is the one we want to visualize at image space. The output is the decoded filter, $df^{L,0}_k$, that we will simplify as $df^L_k$, and which compiles the effect of the CNN architecture under the filter $f^L_k$ in layer $L$.

\section{Discussion} \label{sec:results}

In order to evaluate the proposed procedure, we have computed several visualizations and experiments. They have been performed on the LeNet CNN trained on the MNIST dataset \cite{Lecun98}. The decision to use this dataset is based on the fact that we need to evaluate a new methodology, and we wanted to be able to visually evaluate the emergence of basic local features in a comprehensive level of  complexity. Further experiments are required on more generic dataset but this out of the scope of this paper. MNIST consists of 70,000 gray scale images of size $28\times 28$. Our LeNet\footnote{It is the defalut architecture to train MNIST dataset using matconvnet toolbox \cite{vedaldi15matconvnet}} has 8 layers. The first one is a convolutional layer with a filter bank of 20 single-channel filters of $5\times 5$ size. The second one is a max pooling layer. Third is another convolutional layer with 50 different filters of $5 \times 5$ followed by another max pooling layer. Layer 5 contains a filter bank of $500$ filters of $4\times4$ pixels. Following up in the architecture, we have a ReLU layer followed by another convolutional layer of $10$ filters with one single pixel. At the end, layer 8 applies the softmax operation in order to get the predicted probabilities. All pooling operations are applied to the set of non-overlaping $2 \times 2$ pixel regions.

The first visualization tries to show the idea that a decoded filter is somehow compiling the effects of several layers in the neural architecture. The convolution of a given image with the decoded filter, $df_k^L$ is estimating the output of the filter $k$ at level $L$. A perfect estimation could be understood as a classifier based on a single convolution. In our LeNet the 10 decoded filters in layer $7$ can be seen as 10 classifiers with a single neuron. In figure \ref{fig:figure1} we show the these 10 filters on the left column. Matrix values are giving the value of the maximum activation for all the images in the dataset, top row presents the images holding this maximum activation. Of course, we can not expect a perfect classification  but we can observe on our decoded filters important features correctly characterizing each class, since they are based on the image space. The matrix diagonal is plotting highest activation values for each corresponding filter-image pair. 
\begin{figure*}[t]
	\includegraphics[width=1\linewidth]{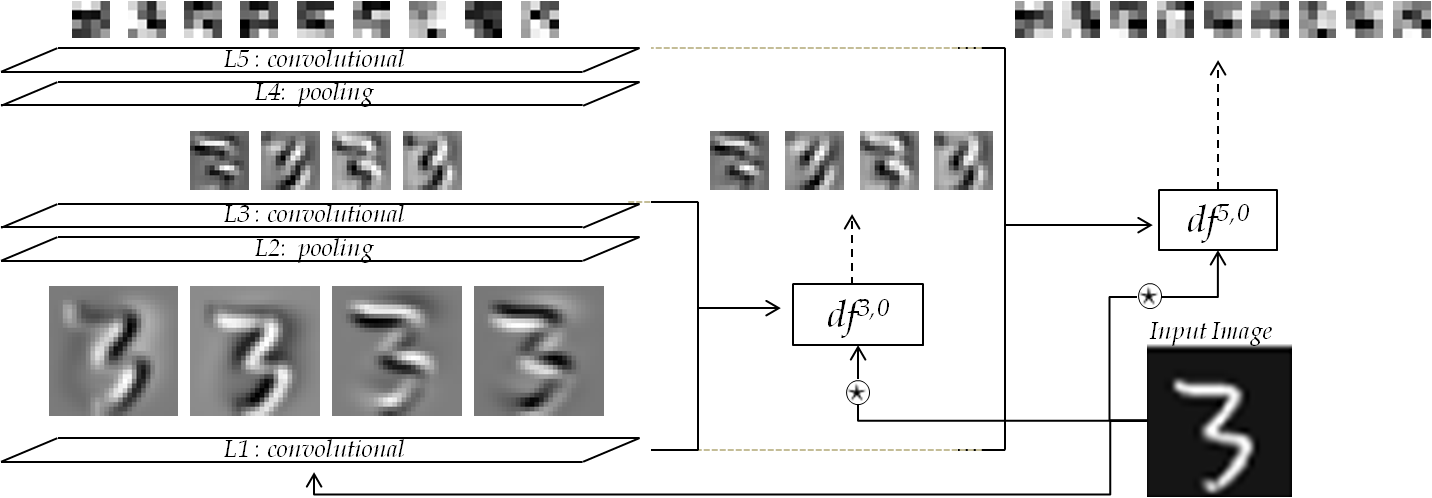}
	\caption{Comparison between image responses obtained through the network (left) and by convolving with decoded filters (right). Responses at different layers show similar activation for both cases, that validate the proposed approach. Obviously the error increase a with the increase in number of encoded layers.}
	\label{fig:compare_responses}			
\end{figure*}

The second visualization pursuit to explore the potential reconstruction power of our approach, in figure \ref{fig:compare_responses} we compare the activations responses at each layer between an image represented with the LeNet at a given level and the convolution of the same image with the corresponding decoded filter bank of the same level.  We show image responses for a class 3 image, trough layers 1, 3 and 5. On the left side we show a subset of image responses on LeNet and on the right side the decoded convolutions. Layer 1 activations of the decoded filter bank are not shown since they are based on the same filter bank. From this plot we can qualitatively conclude that the partial responses built with our method are quite well recovered.

The quantitative evaluation of the previous analysis is based on the mean square error between the CNN response and our reconstruction. The formulation is given in the following equation \ref{eq:error} 

	\begin{equation}
	Error= Mean(||CNN(I,l,k) - (I \star df^l_k) ||^2    )
	\label{eq:error}
	\end{equation}

this error is averaged over all the images in the dataset after a single contrast range normalization for each image. All the computed errors are given in table \ref{tab:error}. As expected, the error increases with the increase of layer depth. All errors are small, independently of the interpolation method used in the unpooling layers.

\begin{table}
	\begin{center}
			\begin{tabular}{| c | c | c | c |} \hline
				& Bicubic &	Nearest &	Bilinear \\
				\hline
				$L3$ &	$0.002 \pm 0.001$ &	 $0.0022 \pm 0.001$ &	 $0.003\pm 0.001 $\\
				$L5$ &	$0.012 \pm 0.005$ &	$0.012 \pm 0.005$ &	$0.013 \pm 0.0055$\\
				$L7$ &	$0.078 \pm 0.056$ &	$0.077 \pm 0.055$ &	$0.081 \pm 0.057$\\
				\hline 		
			\end{tabular}
		\end{center}
	\caption{Mean square error as in equation \ref{eq:error} for layers L3, L5 and L7, and for 3 different unpooling interpolation}\label{tab:error}
\end{table}

\begin{figure*}
	\begin{center}
		\includegraphics[width=0.85\linewidth]{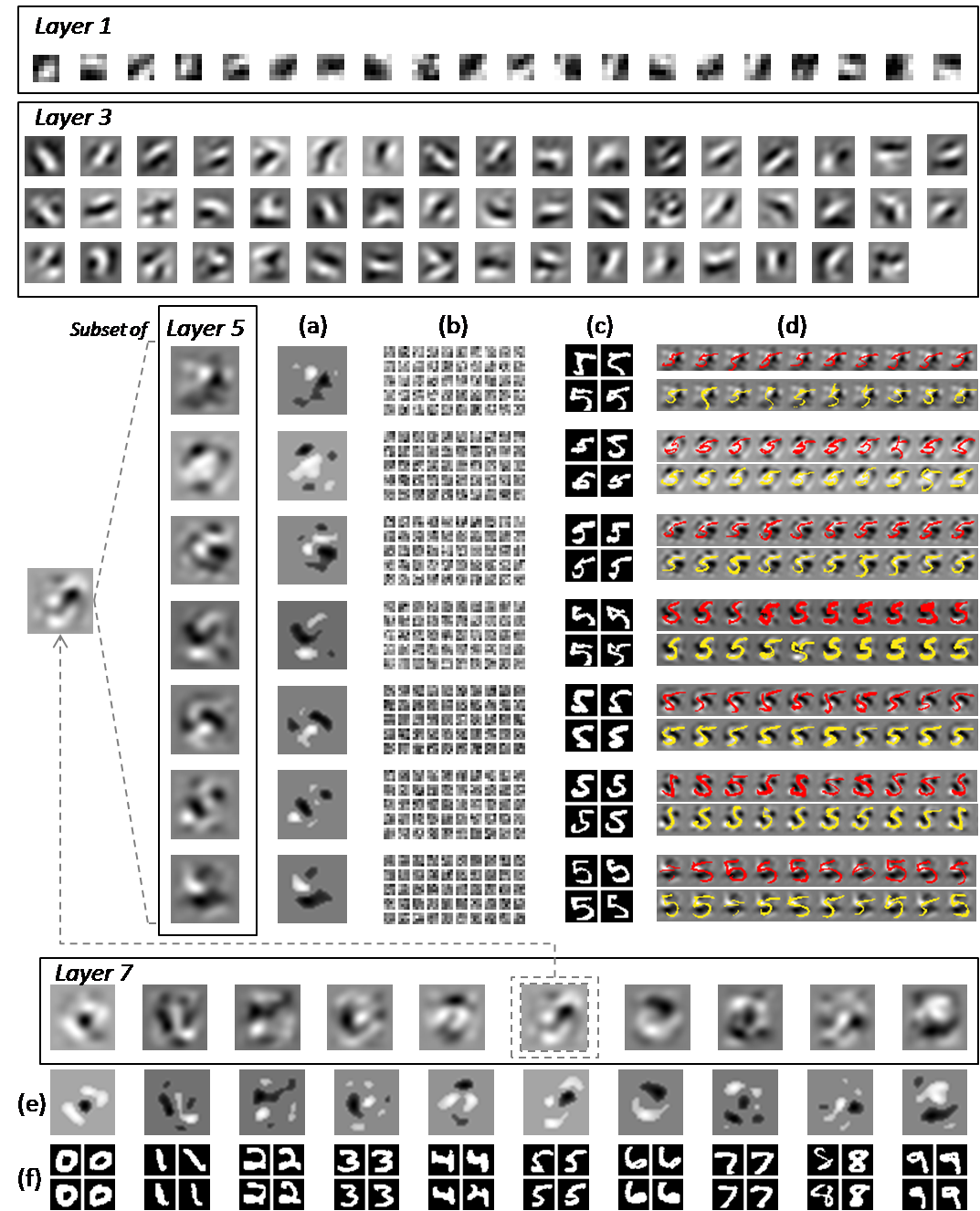}
	\end{center}
	\caption{LeNet decoded filters. (Layer 1) 20 ($5\times 5$) decoded filters which are the original trained filters, most are oriented edge detectors. (Layer 3) 50 ($14 \times 14$) decoded filters increasing complexity with respect to layer 1. They are oriented versions of bars, corners, terminators and crossings. (Layer 5) 7 ($28 \times 28$) decoded filters from a subset of 500, they were selected to show filters sharing high activations for a common class, e.g. 5. Complexity increases with respect to previous layers, representing patterns adapted to different subclasses of 5's. (Layer 7) 10 ($28 \times 28$) decoded filters presenting representative patterns for each class. (a) and (e) Threshold outputs of decoded filters in layers 5 and 7 respectively, they enhance the positive and negative basic patterns. (b) Original trained filters in layer 5 before decoding ($5 \times 5$ and $20$ channels each). (c) and (f) First four images with highest activations for the corresponding filter. (d) Original images for which a maximum activation of the corresponding filter has been obtained with LeNet (in red) and with the convolution with decoded filter (in yellow), shown in first and second rows respectively, both overlap the corresponding decoded filter. All red and yellow fives for a given filter belong to a similar subclass.} \label{fig:all_filters}
\end{figure*}	
		
The third visualization shown by figure \ref{fig:all_filters} where we plot all the LeNet decoded filters for layers 1, 3 and 7 and a subset for layer 5, since the number of filters (500) is too large. The most interesting point is that we clearly can observe an increase in complexity as we move deeper in the net. In the first layers we can observe oriented edge detectors, while we have oriented versions of bars, corners, terminators and crossings in layer 3. Decoded filter for Layer 5 already cover the image size, presenting different versions of the same class, we only show a subset of the 500 filters we have in this layer. A nice observation is that neurons are grouping features based on shape similarity that correlates with recent biological evidences shown by Baldasi \etal \cite{dicarlosimilarity}. We plot all the filters for layer 7, where we can easily see how basic feature of each class are easily perceived.

\section{Conclusion and further work}\label{sec:conclusions}

We have developed a methodology to explore the features learned when trainig a CNN network. We have proposed an algorithm for filter decoding that brings the filter representation from a given layer to the image space. In this way we are able to visualize and therefore understand what is the task of any filter in the hierarchy. 

The main contribution arises from the fact that we have proposed a new way to understand deconvolution. We point out about the concepts of convolution and correlation and the lack of a generic solution for the inverse computation. From these observation we propose the correlation assumption that states that a substitution of the filter shape is a better approximation of the deconvolution step since it is inverting the meaning of activation from one layer to another. 

Our proposal has been evaluated on the intermediate representations of the network filters for the MNIST dataset. We have computed the mean square error with respect to the representation provided for our decoded filter wich is compiling all the levels below the filter. Computed error are small for all the layers. 

This work is presenting a first insight on the filter decoding solution. As future work we should apply the methodology on trained CNNs on natural datasets and we also need a more in depth analysis about the effects of the specific operations in the networks such as relu or other non-linearities.

{\small
	\bibliographystyle{ieee}
	\bibliography{egbib}

\begin{thebibliography}{10}\itemsep=-1pt

\bibitem{Dosovitskiy15a}
T.~B. Alexey~Dosovitskiy, Jost Tobias~Springenberg.
\newblock Learning to generate chairs with convolutional neural networks.
\newblock In {\em CVPR}, 2015.

\bibitem{Aubry15}
M.~Aubry and B.~C. Russell.
\newblock Understanding deep features with computer-generated imagery.
\newblock In {\em ICCV}, 2015.

\bibitem{dicarlosimilarity}
C.~Baldassi, A.~Alemi-Neissi, M.~Pagan, J.~J. {DiCarlo}, R.~Zecchina, and
  D.~Zoccolan.
\newblock Shape similarity, better than semantic membership, accounts for the
  structure of visual object representations in a population of monkey
  inferotemporal neurons.
\newblock {\em PLoS computational biology}, 9, 2013 Aug 2013.

\bibitem{dicarlo14}
C.~F. Cadieu, H.~Hong, D.~L.~K. Yamins, N.~Pinto, D.~Ardila, E.~A. Solomon,
  N.~J. Majaj, and J.~J. {DiCarlo}.
\newblock Deep neural networks rival the representation of primate it cortex
  for core visual object recognition.
\newblock {\em PLoS computational biology}, 10:e1003963, 2014 Dec 2014.

\bibitem{Dosovitskiy15b}
A.~Dosovitskiy and T.~Brox.
\newblock Inverting convolutional networks with convolutional networks.
\newblock 2015.

\bibitem{Kavukcuoglu10}
K.~Kavukcuoglu, P.~Sermanet, Y.-L. Boureau, K.~Gregor, M.~Mathieu, and
  Y.~LeCun.
\newblock Learning convolutional feature hierachies for visual recognition.
\newblock In {\em Advances in Neural Information Processing Systems (NIPS
  2010)}, volume~23, 2010.

\bibitem{Lecun98}
Y.~Lecun, L.~Bottou, Y.~Bengio, and P.~Haffner.
\newblock Gradient-based learning applied to document recognition.
\newblock In {\em Proceedings of the IEEE}, pages 2278--2324, 1998.

\bibitem{Mahendran15}
A.~Mahendran and A.~Vedaldi.
\newblock Understanding deep image representations by inverting them.
\newblock In {\em CVPR}, 2015.

\bibitem{vedaldi15matconvnet}
A.~Vedaldi and K.~Lenc.
\newblock Matconvnet -- convolutional neural networks for matlab.

\bibitem{Yosinski15}
J.~Yosinski, J.~Clune, A.~M. Nguyen, T.~Fuchs, and H.~Lipson.
\newblock Understanding neural networks through deep visualization.
\newblock 2015.

\bibitem{Zeiler14}
M.~D. Zeiler and R.~Fergus.
\newblock {\em Visualizing and Understanding Convolutional Networks}.

\bibitem{Zeiler10}
M.~D. Zeiler, D.~Krishnan, G.~W. Taylor, and R.~Fergus.
\newblock Deconvolutional networks.
\newblock In {\em CVPR}, 2010.

\bibitem{Zeiler11}
M.~D. Zeiler, G.~W. Taylor, and R.~Fergus.
\newblock Adaptive deconvolutional networks for mid and high level feature
  learning.
\newblock In {\em ICCV}, 2011.

\end{thebibliography}
}

\newpage

\part *{Supplementary Material}

\begin{abstract}
	In this supplementary material we provide with some illustrations to reinforce de comprehension of the filter decoding procedure presented in the original paper. We emphasize the idea of the local substitution as a valid estimate of the inverse convolution for convolutional networks. To this end we built very simplistic character images in order to plot the hierarchical representation of image subshapes indpendently of small spatial variations. In this way we strength the role of the filter representation and consequently the meaning of its decoded version. 
\end{abstract}

\section{Filter Decoding on a simpler case}
Filter decoding is a procedure derived from two main ideas. First, we remark that CNNs are based on shape correlation instead of convolution itself. Second,  pixel representing an activation output a a convolutional layer can be understood as a similiarity measurement between the filter shape and the area surrounding that pixel in the inferior layer. Both ideas bring us to define a method to decode CNN trained filters towards their representation at the image space.

In the paper we presented a visualization of decoded filters on a certain CNN (LeNet) trained on MNIST and we could analyze the main features that were learned by the architecture. Here we try to go more in depth on the main ideas of our proposal by showing the effects of our algorithm on simpler images that can help in the understanding of the proposal. 

\begin{figure*}[t]
	\begin{center}
		\includegraphics[width=1\linewidth]{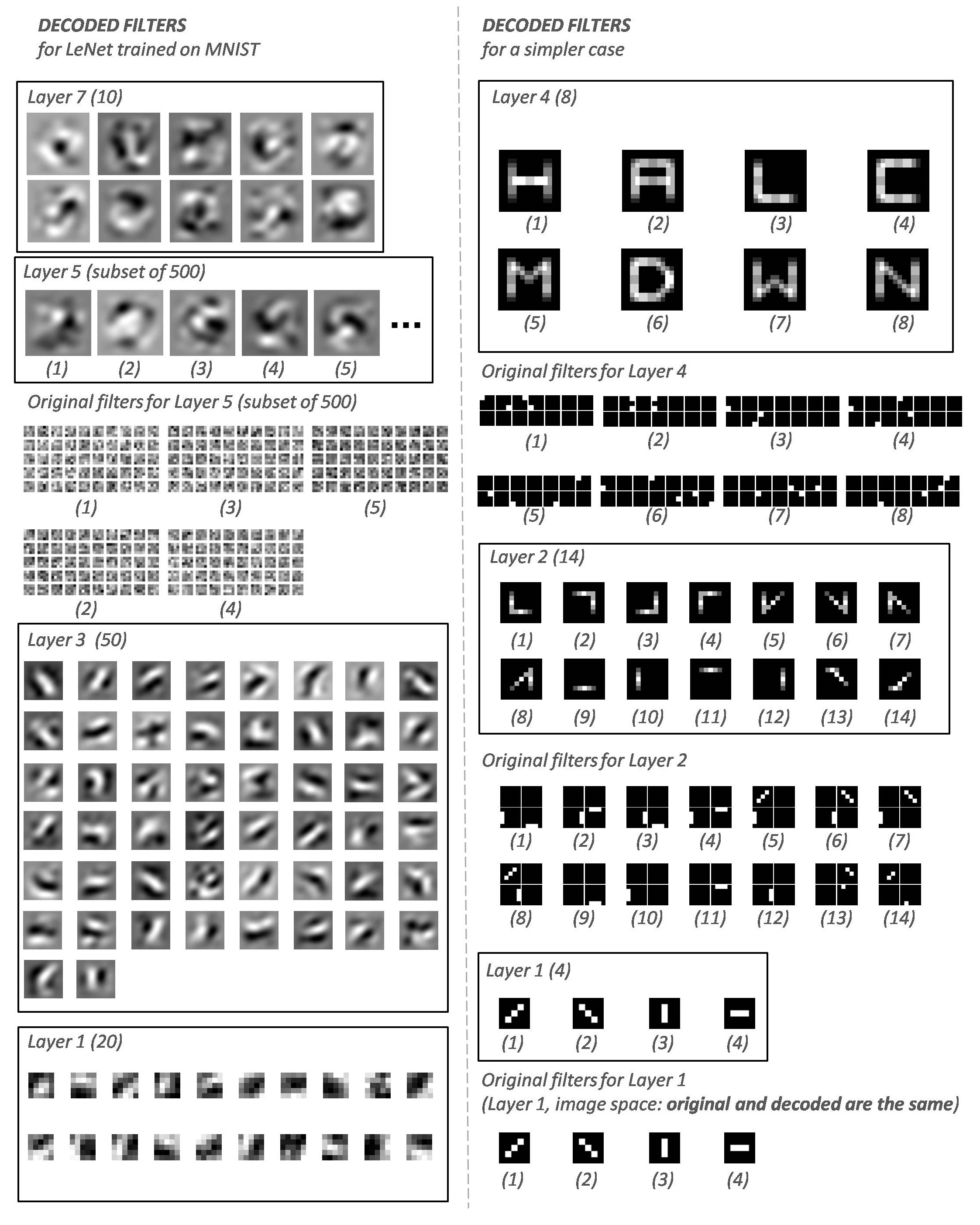}
	\end{center}
	\caption{Examples of DECODED FILTERS. (a) Set or subsets of decoded filter provided by our method on the LeNet trained with MNIST grouped by layers (partially reproduced in figure 4 of the paper). For Layer 5 we also plot some of the original filters.  (b) Full set of decoded filters provided by the proposed method on a basic net built on simple filters. Original filters are also given for each convoltuional layer. }
	\label{fig:figure1_sm}
\end{figure*}

In figure \ref{fig:figure1_sm} we show the family of decoded filter at each layer. On the left side we re-plot the filters obtained on the MNIST dataset on LeNet (figure 4 in the paper). On the right side we plot a new set of decoded filters obtained with the application of the same algorithm on a simpler net. In this case, filters have been specifically built for a naive example that can easily show up the hierarchical representation of complex shapes based on simple intrinsic shapes. The original filters we have defined are those whose decoded version allow to represent specific sub-shapes without any degree of variation, in order to facilitate the comprehension of the defined concepts. This handmade net has been designed with 4 layers. Three convolutional layers, 1, 2 and 4 and one polling layer, 3. The filters are  $4\;@\;(5\times5)$,   $14\;@\;(5\times5)$ and $8\;@\;(3\times3)$, in layers 1, 2 and 4, respectively. 

\begin{figure*}[t]
	\begin{center}
		\includegraphics[width=1\linewidth]{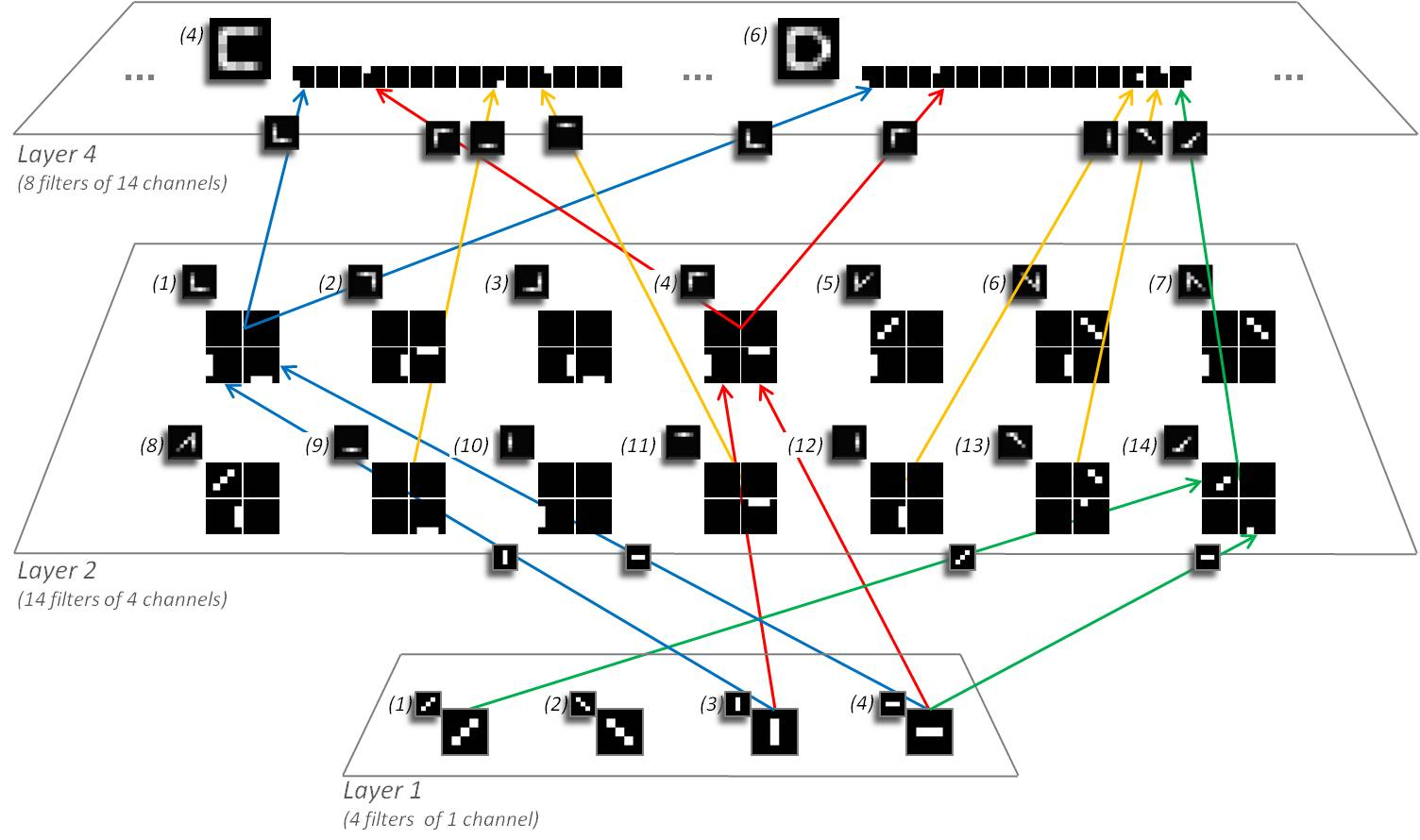}
	\end{center}
	\caption{CNN representation for a subset of neurons. For each neuron we show its filter and its decoded version on the left side.  We only plot those arrows representing  activations of the filter in the inferior layer. }
	\label{fig:figure2_sm}
\end{figure*}

In figure \ref{fig:figure2_sm} we visualize some neuron connections between layers for specific filters. Character \textit{C} is represented by main 4 activation provided the corresponding filters, (1), (4), (9) and (11) in the second layer, that easily explain how the shape is built with the decoded shapes in the inferior layer. Recursively, we can see that decoded subshape associated to filter (1) in layer 2 is provided by the activations of filters (3) and (4) of the first layer. In the same figure, we also show how two different classes share specific subshapes, both representations for characters \textit{C} and  \textit{D}, share a common subshape given by filter (1) of layer 2, which corresponds to the corner located in both cases in the bottom left pixel of layer 4.

\begin{figure*}[t]
	\begin{center}
		\includegraphics[width=1\linewidth]{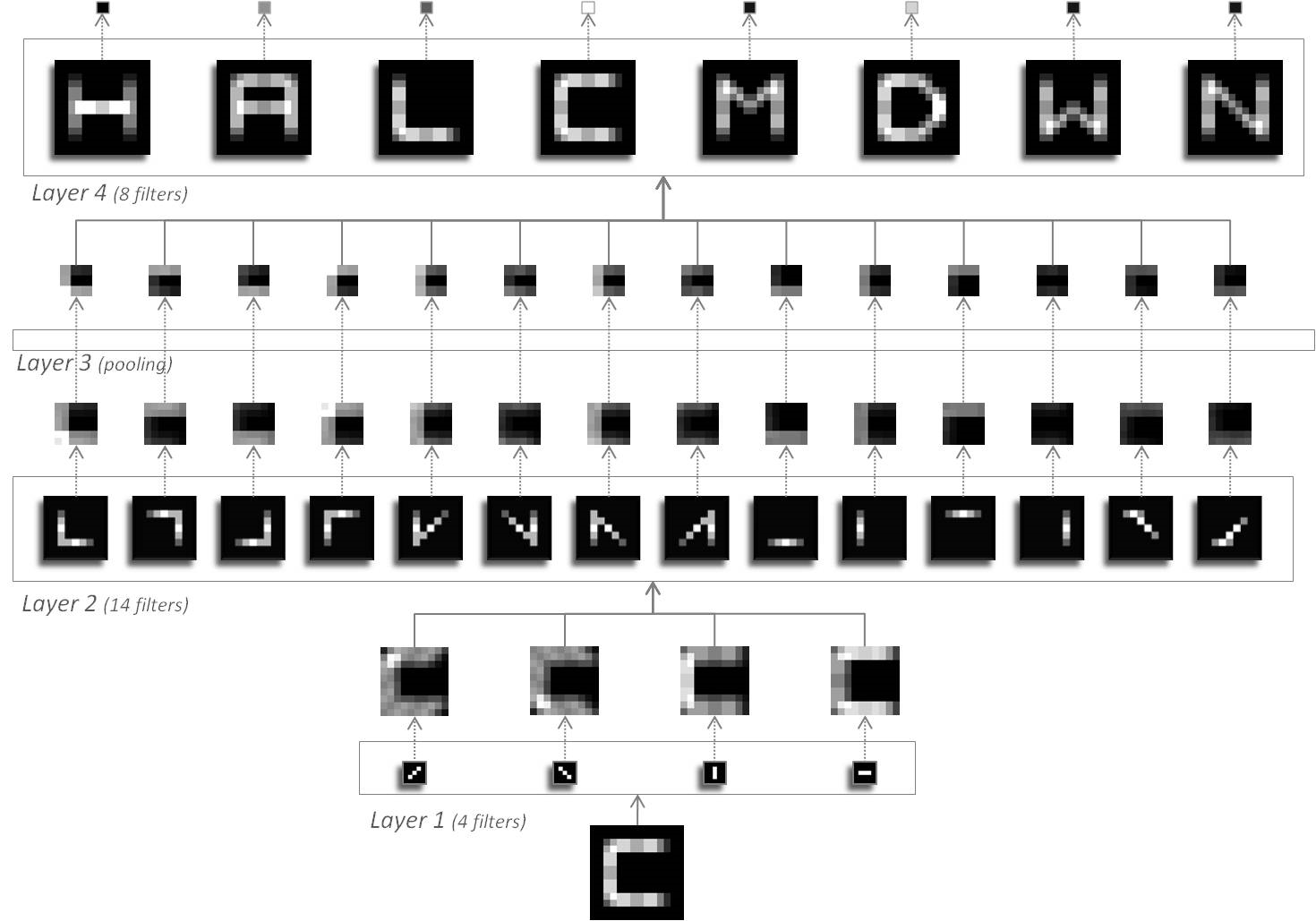}
	\end{center}
	\caption{CNN activations at all layers for a single image}
	\label{fig:figure3_sm}
\end{figure*}

Previous figures help us to understand CNN architecture and the meaning of our decoded filters. As we designed manually the filters of the network, we want to prove if they are consistent with the CNN coding architecture. For this reason, we test our CNN on an image, which is exactly what we have obtained as a decoded filter of a filter at the top convolutional layer. We have tested for all the eight filters of the layer 4, and the results where as we expected: the highest activations at each level are associated with the shape shown by the corresponding decoded filter. Therefore, the analysis of the activations is coherent for all the synthesized characters. Highest activation of a given decoded filtered as being the input of the CNN gives the maximum activation at the corresponding filter at the top layer.  This fact add evidences to the use of the local substitution as the inverse of the convolution (or correlation) in a CNN hierarchical architecture. Figure \ref{fig:figure3_sm} shows all af the activations maps for the character \textit{C}. Highest activations (in white) corresponds with the filter on which the CNN has been applied. Note that this figure shows at each layer the decoded filters instead of the filters itself. This decision has been taken due to space constraints and to help us the comprehension of the activations.

\end{document}